\documentclass[10pt,twocolumn,letterpaper]{article}
\usepackage[accsupp]{axessibility}

\usepackage{iccv}              

\definecolor{iccvblue}{rgb}{0.21,0.49,0.74}
\usepackage[pagebackref,breaklinks,colorlinks,allcolors=iccvblue]{hyperref}
\usepackage{algorithm}
\usepackage{multirow}
\usepackage{makecell}
\usepackage{xcolor}
\usepackage{colortbl}
\usepackage{algpseudocode}
\algtext*{EndIf}

\title{SSVQ: Unleashing the Potential of Vector Quantization with Sign-Splitting}

\author{
   Shuaiting Li$^{1,2}$\thanks{Work done during an internship at vivo Mobile Communication.}, Juncan Deng$^{1,2}$\footnote[1]{} , Chengxuan Wang$^{1}$, Kedong Xu$^{2}$, Rongtao Deng$^{2}$,
   Hong Gu$^{2}$, \\ 
   Haibin Shen$^{1}$\thanks{Corresponding Authors}, Kejie Huang$^{1}$\footnote[2]{}\\ 
   $^{1}$Zhejiang University  \quad \quad
   $^{2}$vivo Mobile Communication Co., Ltd \quad \quad \\
  \small\texttt{\{list,dengjuncan,wangchengxuan,shen\_hb,huangkejie\}@zju.edu.cn}\\
  \small\texttt{\{xukedong, dengrongtao, guhong\}@vivo.com} \\
}

\begin{document}
\maketitle
\begin{abstract}
Vector Quantization (VQ) has emerged as a prominent weight compression technique, showcasing substantially lower quantization errors than uniform quantization across diverse models, particularly in extreme compression scenarios. However, its efficacy during fine-tuning is limited by the constraint of the compression format, where weight vectors assigned to the same codeword are restricted to updates in the same direction. Consequently, many quantized weights are compelled to move in directions contrary to their local gradient information. To mitigate this issue, we introduce a novel VQ paradigm, Sign-Splitting VQ (SSVQ), which decouples the sign bit of weights from the codebook. Our approach involves extracting the sign bits of uncompressed weights and performing clustering and compression on all-positive weights. We then introduce latent variables for the sign bit and jointly optimize both the signs and the codebook. Additionally, we implement a progressive freezing strategy for the learnable sign to ensure training stability. Extensive experiments on various modern models and tasks demonstrate that SSVQ achieves a significantly superior compression-accuracy trade-off compared to conventional VQ. Furthermore, we validate our algorithm on a hardware accelerator, showing that SSVQ achieves a 3$\times$ speedup over the 8-bit compressed model by reducing memory access. Our code is available at \url{https://github.com/list0830/SSVQ}.
\end{abstract}

\section{Introduction}
Deep Neural Networks (DNNs) have demonstrated remarkable success across various computer vision tasks, including image classification~\cite{sandler2018mobilenetv2, liu2022convnet, tan2019efficientnet, touvron2021training}, object detection~\cite{wang2024yolov9}, segmentation~\cite{chang2023yolor, chen2017rethinking}, and generation~\cite{rombach2022high, hu2025dynamicid}. The proliferation of edge intelligence has driven significant interest in deploying DNNs to edge devices, particularly for real-time and privacy-sensitive applications. However, the rapid growth of model parameters often exceeds the capacity of accelerator on-chip SRAMs~\cite{pullini2019mr, rossi20214}, necessitating frequent off-chip DRAM accesses that substantially increase latency and power consumption.

Achieving low-power computing requires hardware-friendly compression algorithms and model optimization. However, current compression methods struggle to balance accuracy and compression ratios. For instance, 1-bit quantization reduces the model size by 32$\times$ but suffers significant accuracy degradation.
While unstructured pruning~\cite{han2015deep,louizos2017learning,renda2020comparing} requires storing indices of unpruned weights, structured pruning~\cite{li2016pruning, molchanov2019importance} rarely exceeds 50\% pruning rates. Vector Quantization~\cite{gong2014compressing,wu2016quantized,Martinez_2021_CVPR,son2018clustering,stock2019and, chen2020towards} (VQ), by grouping and clustering weights, offers a promising solution to achieve higher compression ratios with hardware efficiency. VQ demonstrates a significantly smaller quantization error than that of uniform quantization, particularly under extreme compression scenarios.

Fine-tuning is always essential for extremely low-bit quantization. Although vector quantization has significantly smaller quantization errors compared to uniform quantization, its potential during fine-tuning is limited.
In vector quantization, to maintain the compressed format, only the codebook can be fine-tuned rather than the weights themselves. This means that all weight vectors clustered to the same codeword can only be updated in the same direction, greatly limiting the fine-tuning potential. Consequently, many weight subvectors are updated towards suboptimal values, which ultimately degrades model accuracy.

To fully harness the potential of vector quantization during fine-tuning, we introduce a novel VQ paradigm called Sign-Splitting Vector Quantization (SSVQ), which empowers each quantized weight to update independently in alignment with its gradient direction. Our approach begins by extracting the sign bits of the weights as a 1-bit mask and clustering the absolute values of the weights. This strategy significantly reduces the number of codewords needed to achieve comparable clustering performance. Next, we introduce implicit variables to learn the sign bits, effectively decoupling the update direction of the sign bits from that of the codebook. Finally, to address the challenge of sign bit oscillations during training, we propose an enhanced iterative freezing mechanism that stabilizes the learning process. This framework not only improves flexibility but also ensures robust and efficient fine-tuning of quantized models.

Experiments demonstrate that our SSVQ achieves a significantly better accuracy-compression trade-off compared to conventional VQ and UQ across a wide range of compression scenarios. Furthermore, we built a hardware simulator supporting SSVQ, which achieves 3$\times$ inference speedup compared to 8-bit compressed models.

The main contributions of the paper are summarized as follows:
\begin{itemize}
\item We introduce a novel VQ paradigm, Sign-Splitting VQ, to boost accuracy gains during fine-tuning.
\item We propose an enhanced freezing strategy to ensure stability during training. 
\item We build a cycle-accurate hardware simulator to validate the speedup gains of our SSVQ.
\end{itemize}
\section{Related Work}
To address the increasing demand for models capable of delivering high performance under memory limitations, it is crucial to focus on both designing parameter-efficient networks and developing model compression techniques.

\textbf{Parameter-efficient Networks.}
In recent years, the design of parameter-efficient neural networks has gained significant attention due to the increasing demand for models that perform well under resource constraints. CNNs have seen substantial advancements with architectures like MobileNets~\cite{qin2024mobilenetv4, howard2019searching, sandler2018mobilenetv2} and EfficientNets~\cite{tan2019efficientnet}, which employ techniques such as depthwise separable convolutions and neural architecture search. InceptionNeXt~\cite{yu2024inceptionnext} integrates multi-scale Inception-like modules to speed up large-kernel-based models. On the other hand, transformers have revolutionized natural language processing and are being adapted for vision tasks. EfficientViT~\cite{liu2023efficientvit} and MobileViT~\cite{mehta2021mobilevit} introduce self-attention approximations that maintain linear complexity with minor impacts on accuracy. Additionally, hybrid architectures that combine the strengths of CNNs and transformers, such as MobileFormer~\cite{chen2022mobile}, FastViT~\cite{vasu2023fastvit}, have emerged as powerful solutions.

\textbf{Vector Quantization.}
Vector quantization compresses weights by utilizing a codebook and a list of assignments, rather than storing the weights directly. The reconstruction of the compressed weight is achieved by looking up the corresponding codeword in the codebook based on the assignment. BGD~\cite{stock2019and} utilizes product quantization~\cite{jegou2010product} and integrates activation into clustering to minimize reconstruction error. PQF~\cite{Martinez_2021_CVPR} concentrates on seeking optimal permutations to enhance compression efficiency. DKM~\cite{cho2022dkm} frames k-means clustering as an attention problem, facilitating the joint optimization of DNN parameters and clustering centroids. Furthermore, MVQ~\cite{li2025mvq} introduces a masked k-means procedure that employs pruning to eliminate unimportant weights, thereby enabling a more accurate approximation of significant weights.

\begin{figure*}
    \centering
    \includegraphics[width=0.9\linewidth]{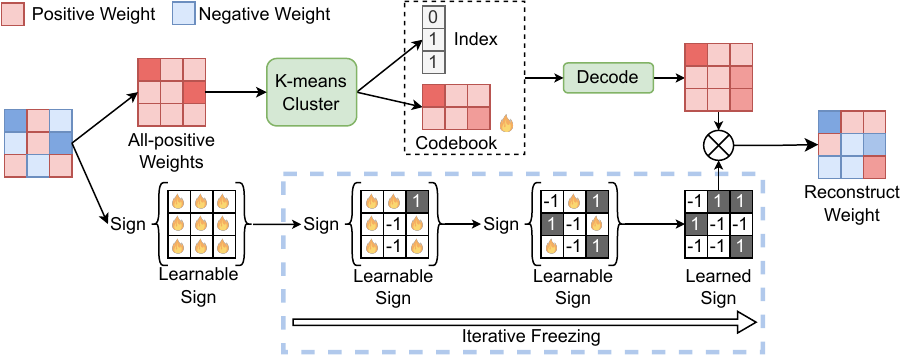}
    \caption{Overview of our Sign-Splitting Vector Quantization (SSVQ) algorithm. Firstly, the sign bit is extracted and k-means clustering is applied to all-positive weights. Secondly, the fixed sign is replaced by the latent parameter's sign function, which is updated based on STE grad approximation. Finally, an improved iterative freezing strategy is employed on the learnable sign to prevent oscillations from destabilizing training. The final storage consists of codebooks, assignments, and sign masks, which are used to reconstruct weights.}
    \label{overview}
    \captionsetup{skip=2pt}
\end{figure*}

\section{Preliminaries}
Let $W \in \mathbb{R}^{O \times I}$ denote the weight matrix of a linear layer. We first partition $W$ into $N = \frac{O \times I}{d}$ subvectors $\{\mathbf{w}_n\}_{n=1}^N$, where each subvector $\mathbf{w}_n \in \mathbb{R}^d$. For simplicity, we can reshape and view $W$ as:
\begin{equation}
    W \Leftrightarrow \{\mathbf{w}_1, \mathbf{w}_2, \dots, \mathbf{w}_N\}.
\end{equation}

\subsection{Codebook Initialization}
In vector quantization, a codebook $\mathcal{C} = \{\mathbf{c}_1, \mathbf{c}_2, \dots, \mathbf{c}_K\}$ is maintained, containing $K$ codewords, where each $\mathbf{c}_k \in \mathbb{R}^d$. The mapping is defined by an assignment function $A: \{1, \dots, N\} \to \{1, \dots, K\}$. The quantized weight tensor $W_q$ is reconstructed by replacing each subvector $\mathbf{w}_n$ with its assigned codeword $\mathbf{c}_{A(n)}$:
\begin{equation}
    W_q \Leftrightarrow \{\mathbf{c}_{A(1)}, \mathbf{c}_{A(2)}, \dots, \mathbf{c}_{A(N)}\}.
    \label{eq:quantized_reconstruction}
\end{equation}
To minimize the quantization error $\|W - W_q\|^2$, the standard approach is to use the \textbf{k-means}~\cite{lloyd1982least, macqueen1967some} algorithm.

\subsection{Codebook Finetuning}
To recover model accuracy after quantization, the codebook $\mathcal{C}$ is fine-tuned. Let $L$ be the network's loss function. For a specific codeword $\mathbf{c}_k \in \mathcal{C}$, let $I_k = \{n \mid A(n)=k\}$ be the set of indices of subvectors assigned to it. The gradient for $\mathbf{c}_k$ is typically approximated by accumulating the gradients of its assigned subvectors, and gradient-based learning is performed as follows :
\begin{equation}
    \mathbf{c}
    _{k}\leftarrow \mathbf{c}_k - O( \sum_{n \in I_k} \frac{\partial L}{\partial \mathbf{w}_n}, \theta)
\end{equation}
where $O(,)$ is the optimizer with hyperparameter $\theta$.

\section{Methodology}
\subsection{Training limitation in Vector Quantization}
\begin{figure}[h]
    \centering
    \captionsetup{skip=4pt}
    \includegraphics[width=0.95\linewidth]{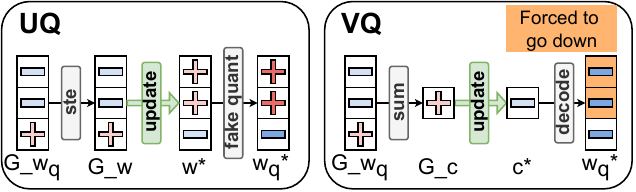}
    \caption{Comparison between uniform quantization and vector quantization on how the quantized weight is updated. We assume a simple update rule, e.g., $w* \leftarrow w - G_{w}$ and $c* \leftarrow c - G_{c}$.}
    \label{UQ_VQ}
    \vspace{-0.2cm}
\end{figure}
We begin by analyzing the differences between uniform and vector quantization in quantization-aware training. As illustrated in Fig.~\ref{UQ_VQ} (left), UQ maintains trainable float point weights $w$, and the gradients $G_{w_q}$ from quantized weights are propagated to $G_{w}$ through the Straight-Through Estimator. The updated $w^{*}$ are then fake quantized to obtain the updated $w_{q}^{*}$, allowing each quantized weight to be updated according to its gradient direction. 

In contrast, VQ's compressed format restricts updates to codewords only. As shown in Fig.~\ref{UQ_VQ} (right), the gradient of a codeword $G_c$ is the sum (in pytorch implementation) of the gradients of the weight vectors clustered to it. This forces all quantized weights within a cluster to share the same update direction. More critically, gradient dominance frequently occurs, where a few strong gradients dictate the update for the entire cluster - as visualized in the orange box of Fig.~\ref{UQ_VQ}, where the bottom weight's gradient forces the upper weights to move downward. 

To verify this, we conduct an empirical analysis on MobileNet-V2 (with $d=4, k=64$).
For each codeword, we compute two gradient statistics from its clustered subvectors: (1) the sum of the top 5-10\% gradients with the largest magnitudes, and (2) the sum of the bottom 50-60\% gradients. Tab.~\ref{emprical study} presents the cosine similarity between gradient subsets and the final codeword gradient. The high correlation with top gradients and low correlation with bottom gradients confirm that a minority of high-magnitude gradients dominate the update, often forcing the majority of weights into suboptimal positions.
This fundamental limitation directly impacts fine-tuning effectiveness, as evidenced by the constrained accuracy improvements brought by training codewords shown in Tab.~\ref{emprical study2}.
\begin{table}[t]
    \centering
    \renewcommand{\arraystretch}{0.8}
    \begin{tabular}{c|cc|cc}
    \toprule
    \multirow{2}{*}{Subset grad} & \multicolumn{2}{c|}{Top grad} & \multicolumn{2}{c}{Bottom grad}\\
    & 5\% & 10\% & 60\% & 50\% \\
    \midrule
    Cosine Similarity & 0.66 & 0.74 & 0.26 & 0.19\\
    \bottomrule
    \end{tabular}
    \captionsetup{skip=4pt}
    \caption{Empirical study on MobileNet-V2 model with conventional VQ. We report the cosine similarity between different subset grad and the codeword grad, confirming gradient dominance.}
    \label{emprical study}
\end{table}

\begin{table}[t]
    \centering
    \captionsetup{skip=4pt}
    \renewcommand{\arraystretch}{0.8}
    \begin{tabular}{c|c|c|c|c}
    \toprule
    Train Param & None & BN & BN+FC & Full \\
    \midrule
    Top-1 Acc & 0.1 & 40.46 & 50.01 & 55.27\\
    \bottomrule
    \end{tabular}
    \caption{Fine-tuning analysis of conventional VQ on MobileNet-V2: Top-1 ImageNet accuracy after 3 training epochs. Layer annotations: BN (batch normalization layers), FC (fully connected layers), Full (codebooks + BN + FC). Results indicate limited accuracy gains from codebook training.}
    \label{emprical study2}
    \vspace{-0.4cm}
\end{table}

\subsection{Sign-Splitting Vector Quantization}
To unleash the potential of vector quantization during the fine-tuning process, we need a new compression paradigm that enables each quantized weight to update in its own gradient direction. To this end, we propose Sign-Splitting Vector Quantization (SSVQ). First, we decouple the weights $W$ into their signs $S$ and absolute values $|W|$. Then, k-means clustering is applied only to the absolute values.
\begin{equation}
    \mathcal{C}, A = \operatorname{k-means}(|W|, K), \quad S = \operatorname{sign}(W)
    \label{ap_kmeans}
\end{equation}
where $S \in \{-1, 1\}^{O \times I}$ is the sign mask. The quantized weight $W_q$ is reconstructed by combining the clustered magnitudes with their signs:
\begin{equation}
    W_q = \mathcal{C}[A] \circ S
\end{equation}

To allow for independent update directions, we make the sign mask learnable. We introduce a continuous, learnable variable $L_s$ initialized as $L_s \leftarrow \alpha \cdot W$ where $\alpha$ is a constant. The forward pass then becomes:
\begin{equation}
    W_q = \mathcal{C}[A] \circ \operatorname{sign}(L_s).
\end{equation}

During backpropagation, we jointly learn the codebook $\mathcal{C}$ and the sign parameter $Ls$. We employ the Straight Through Estimator to approximate the gradients for the non-differentiable sign function. Specifically, the gradient of $Ls$ is computed as
\begin{equation}
    g_{Ls} = \frac{\nabla L}{\nabla W_q} \circ \frac{\nabla W_q}{\nabla sign(Ls)} \circ \frac{\nabla sign(Ls)}{\nabla Ls} \approx \frac{\nabla L}{\nabla W_q} \circ c 
\end{equation}

While the direct application of STE to approximate gradients lacks precision, the gradient of $Ls$ incorporates the magnitude of the corresponding codeword. This inclusion, to some extent, helps stabilize training, as supported by the findings in \cite{liu2018bi}.
\begin{figure}[h]
    \centering
    \captionsetup{skip=2pt}
    \includegraphics[width=0.95\linewidth]{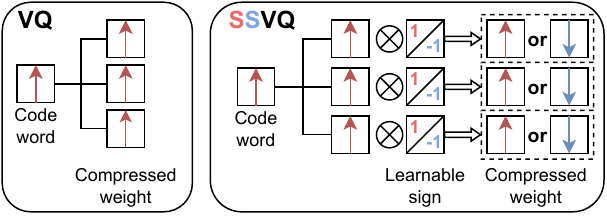}
    \caption{Comparison between VQ and SSVQ. SSVQ enables a quantized weight to update in a different direction with the codeword using learnable signs, which VQ can't.}
    \label{compare}
\end{figure}

 Although the sign mask requires 1 bit of storage, leveraging the inherent properties of clustering allows the same clustering error for all-positive values to be achieved with significantly fewer centroids.

Therefore, SSVQ enables the expansion of the quantization dimension into higher-dimensional spaces while utilizing smaller codebooks. 
The inherent properties of clustering effectively offset the 1-bit storage overhead introduced by the sign representation. At the same compression ratio, SSVQ has a similar clustering error compared to VQ. A detailed analysis is provided in the Supple. Material (Sec.~\ref{sec:initial error}).
\subsection{Enhanced Iterative Freezing of Learnable Signs}
\begin{figure}[h]
    \centering
    \includegraphics[width=0.95\linewidth]{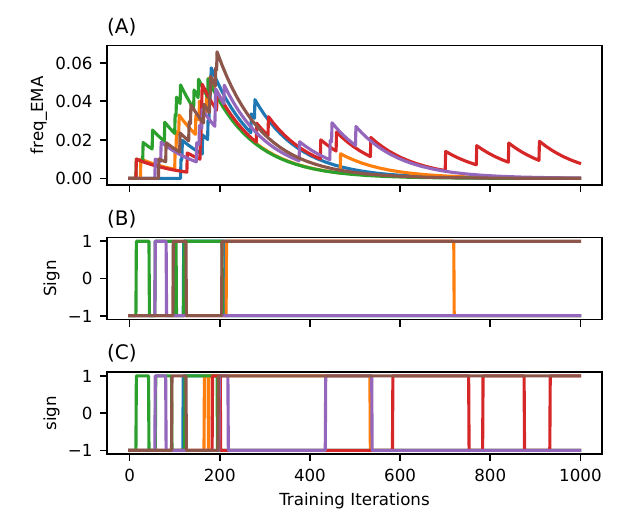}
    \captionsetup{skip=2pt}
    \caption{Oscillation patterns in a 1.5-bit quantized MobileNet-v2 during fine-tuning: (A) freq-EMA oscillation, (B) frozen results oscillation based on majority sign voting, and (C) frozen results oscillation based on sign-EMA, sampled from six random positions in the final convolutional layer.}
    \label{ema}
\end{figure}

\begin{algorithm}[h]
    \small
    \renewcommand{\algorithmicrequire}{\textbf{Input:}}  
    \renewcommand{\algorithmicensure}{\textbf{Output:}}  
    \caption{Iterative Freezing of Learnable Signs}
    \begin{algorithmic}[1]
        \Require Freeze iteration $Fi$, Momentum $m$
        \Require Freeze threshold cosine scheduler $T_{cos}$
        \Require Init: $f \gets 0$, $p_{c} \gets 0$, $n_{c} \gets 0$
        \For{$t = 1$ to $T$}  
            \State Update $Ls$ using $g_{Ls}^{t} = \frac{\nabla L}{\nabla W_q} \circ c$
            \State $s^{t} = \operatorname{sign}(Ls^{t}) \neq \operatorname{sign}(Ls^{t-1})$
            \State $f^{t} = f^{t-1} \times m + s^{t} \times (1 - m)$
            \If{$f^{t} \neq 0$}
                \State $p_{c} = p_{c} + (\operatorname{sign}(Ls^{t}) > 0)$
                \State $n_{c} = n_{c} + (\operatorname{sign}(Ls^{t}) < 0)$
            \EndIf
            \If{$t \% Fi = 0$}
                \If{$f^{t} > T_{cos}(t)$}
                    \If{$p_{c} > n_{c}$}
                        \State $Ls \gets +\infty$ \Comment{sign(Ls) remains positive}
                    \Else
                        \State $Ls \gets -\infty$ \Comment{sign(Ls) remains negative}
                    \EndIf
                \EndIf
            \EndIf
        \EndFor
    \end{algorithmic}
    \label{freezing}
\end{algorithm}

When a sign flips, the value of the quantized weight changes by a magnitude of $2|\mathbf{c}|$. This large, discrete change causes the learnable signs to oscillate, as both $+|\mathbf{c}|$ and $-|\mathbf{c}|$ may be suboptimal approximations. A similar issue was observed in uniform quantization-aware training~\cite{nagel2022overcoming}, where weights oscillate between quantization grid points. They proposed using two Exponential Moving Averages (EMAs) to separately track the oscillation frequency and the frozen results. Inspired by ~\cite{nagel2022overcoming}, we further analyze the sign dynamics in SSVQ and identify two key phenomena:
\begin{itemize} 
\item As shown in Fig.~\ref{ema}(A), the majority of freq-EMA initially show substantial fluctuations, which gradually decrease as training progresses. If the learnable sign is frozen too frequently (for example, by evaluating the freezing in every iteration as suggested in ~\cite{nagel2022overcoming}), there is a risk of capturing and freezing unstable results prematurely.

\item As shown in Fig.~\ref{ema}(C), the frozen result based on sign-EMA also exhibits frequent oscillations. Since the sign is binary, a simpler majority vote is more robust. Fig.~\ref{ema}(B) shows that basing the frozen sign on a simple majority count of positive vs. negative signs results in a much more stable outcome.
\end{itemize}

Based on these observations, we propose an enhanced iterative freezing strategy for learnable signs, as summarized in Algorithm~\ref{freezing}. In each iteration, we update freq-EMA $f$ and record the positive $p_{c}$ and negative counts $n_{c}$ for learnable signs that have undergone sign changes. For every $Fi$ iteration, we identify signs whose oscillation frequency exceeds a decaying threshold and freeze them permanently to their majority-voted sign. We apply a cosine schedule to the freezing threshold, similar to~\cite{nagel2022overcoming}.

\section{Experiments}
\begin{figure*}
    \centering
    \captionsetup{skip=2pt}
    \includegraphics[width=0.95\linewidth]{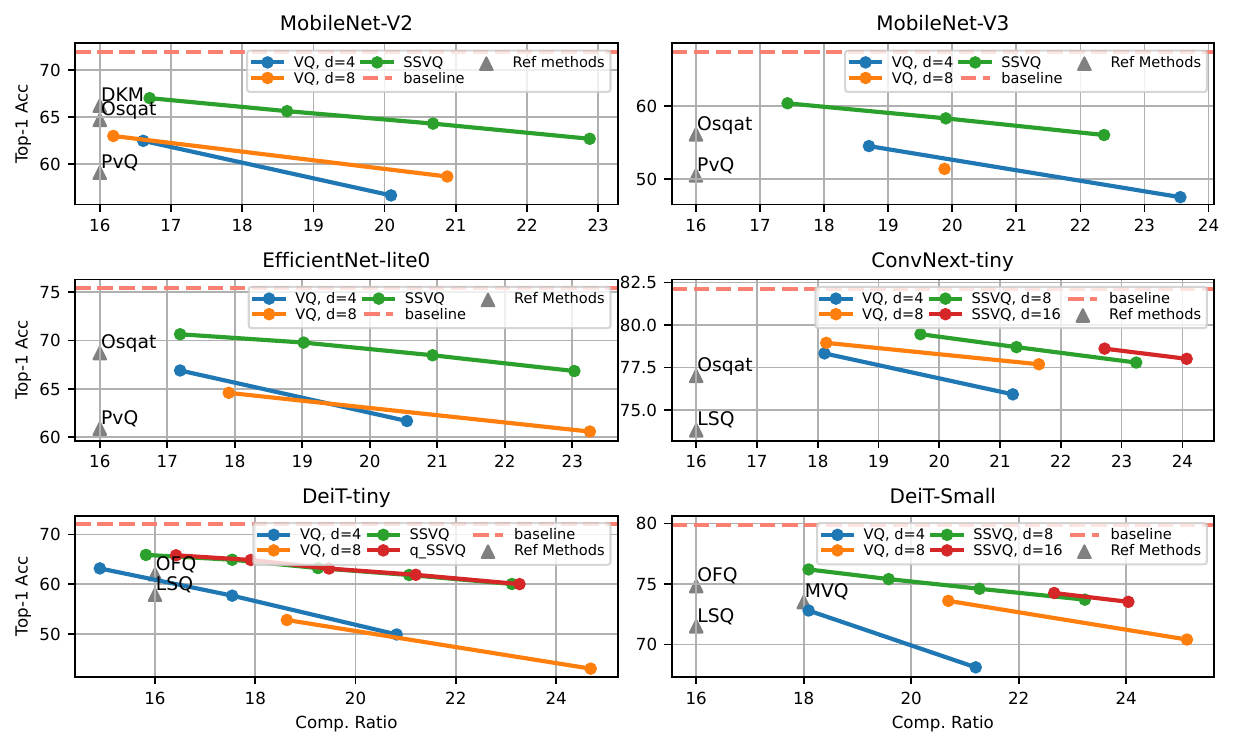}
    \caption{Quantization results of classification on ImageNet dataset. We present the accuracy-compression trade-off curves for both VQ and SSVQ, with competing methods annotated as discrete points for comparative analysis. The full-precision results are marked as \textcolor{pink}{'baseline'}.}
    \label{classify}
    \vspace{-0.1cm}
\end{figure*}
We systematically validate the effectiveness of SSVQ through two complementary analyses. First, extensive experiments across diverse architectures and tasks demonstrate consistent improvements of SSVQ over conventional VQ and existing methods. Second, ablation studies quantify the contributions of our core innovations.
\subsection{Storage Analysis}
The compression ratio (CR) is denoted as the key metric. Denote $b_f$ as the bit-width of full-precision weight. Assuming that the codebook is quantized to $q_c$ bits, the CR of $W \in \mathbb{R}^{O \times I}$ after employing conventional VQ is calculated as:
\begin{equation}
    \text{CR}_{\text{VQ}} = \frac{O \cdot I \cdot b_f}{b_a + b_c} = \frac{d_1 \cdot b_f}{\lceil\log_{2}(k_1)\rceil + \frac{k_1 \cdot d_1^2 \cdot q_c}{O \cdot I}}
\end{equation}
where $b_a$ and $b_c$ are the storage costs of assignments and codebooks, respectively. 

As for SSVQ, we have to store the 1-bit sign mask $b_s$. The CR is thus calculated as:
\begin{equation}
    \text{CR}_{\text{SSVQ}} = \frac{O \cdot I \cdot b_f}{b_a + b_c + b_s} = \frac{d_2 \cdot b_f}{d_2 + \lceil\log_{2}(k_2)\rceil + \frac{k_2 \cdot d_2^2 \cdot q_c}{O \cdot I}}
\end{equation}

\textbf{We set different $k$ and $d$ to ensure comparable average bit-width between VQ and SSVQ. Since clustering on all-positive weights is much easier, SSVQ requires much fewer index bits (i.e. smaller $k$ and larger $d$ ).} We provide detailed $k \& d$ settings of all experiments in Sec.~\ref{sec:quant settings} in the Supplementary Material.

\subsection{Experimental setup}
\textbf{Implementation details.}
We employ the AdamW optimizer with a cosine annealing learning rate schedule across all tasks. To ensure reproducible comparative analysis, we maintain a consistent 10-epoch training regimen without distillation for both classification and detection tasks. Uniform quantization benchmarks are reproduced under the standard `w2` configuration. Image generation experiments follow the established pipeline in ~\cite{li2024efficiency}, while semantic segmentation builds upon ~\cite{bubbliiiing_deeplabv3_plus_pytorch}. The sign variation ratio is modulated through either the learning rate or the initialization parameter $\alpha$. Detailed hyperparameter specifications are available in the supplementary material.

\textbf{Comparison methods.}
Given the limited availability of VQ implementations validated on diverse modern vision architectures (many previous VQ works~\cite{stock2019and, Martinez_2021_CVPR, liu2023hyperspherical, cho2022dkm, li2025mvq, zhu2023learning} still focus greatly on ResNets), we systematically implemented conventional VQ baseline with identical training protocols to SSVQ for each task category, establishing comprehensive baseline evaluations. For comparative analysis, we also present results from contemporary VQ methods (MVQ~\cite{li2025mvq}, HQ~\cite{liu2023hyperspherical}, DKM~\cite{cho2022dkm}) and state-of-the-art UQ approaches (LSQ~\cite{esser2019learned}, PvQ~\cite{kuzmin2023pruning}, Osqat~\cite{nagel2022overcoming}, OFQ~\cite{liu2023oscillation}, EMF~\cite{li2024efficiency}, PCR~\cite{tang2024post}, Quest~\cite{wang2024quest}, MPQ-DM~\cite{feng2024mpq}).

\subsection{Performance Evaluation}
We thoroughly assess the effectiveness of SSVQ across various vision tasks including image classification, object detection, segmentation, and text-to-image generation.
Notably, 'C.R' consistently denotes the compression ratio of compressed layers. 

\textbf{Results for classification.}
Our evaluation begins with ImageNet classification benchmarks, with quantization performance summarized in Fig.~\ref{classify}. SSVQ achieves significant accuracy gains over conventional VQ and UQ, particularly under extreme compression. For example, SSVQ improves the accuracy of DeiT-tiny by 12\% absolute at a 21 $\times$ compression ratio compared to baseline VQ implementations.
For CNN architectures, where VQ typically compresses fully-connected (FC) layers while UQ maintains FC layers at 8-bit/FP16 precision, we specifically include comparative FC layer compression results in Tab.~\ref{fc}, demonstrating SSVQ's superior accuracy over SOTA VQ methods.
\begin{table}[h]
    \centering
    \captionsetup{skip=2pt}
    \renewcommand{\arraystretch}{0.8}
    \setlength{\tabcolsep}{0.7mm}
    \begin{tabular}{c|ccc|cc}
    \toprule
    Models& \multicolumn{3}{c|}{MobileNet-v2 (71.9)} & \multicolumn{2}{c}{EfficientNet-lite (75.4)} \\
    \midrule
    Methods & MVQ & HQ & SSVQ & MVQ & SSVQ \\
    \midrule
    C.R & 16$\times$ & 20$\times$ & 20/16$\times$  & 16$\times$ & \textbf{18$\times$}\\
    \midrule
    Acc &65.1 & 58.8 & \textbf{63.1/65.9} & 68.2 & \textbf{69.5} \\
    \bottomrule
    
    \end{tabular}
    \caption{Additional quantization results on CNNs which also compress fully-connect layers.}
    \label{fc}
\end{table}

\textbf{Results for Semantic Segmentation.}
The quantization results for semantic segmentation tasks on the VOC dataset are presented in Tab.~\ref{deeplab}. Our SSVQ outperforms MVQ by 1.2\% mIoU at a 19$\times$ compression ratio.
\begin{table}[h]
    \centering
    \captionsetup{skip=2pt}
    \renewcommand{\arraystretch}{0.8}
    \setlength{\tabcolsep}{1.2mm}
    \begin{tabular}{c|c|c|c|c|c}
    \toprule
    Method & FP & VQ & PvQ~\cite{li2016pruning} & MVQ~\cite{li2025mvq}  & SSVQ \\
    \midrule
    C.R & - & 17.5$\times$ & 16 $\times$ & 19$\times$ & \textbf{19}$\times$\\
    \midrule
    mIoU & 72.9 & 65.7 & 19.2 & 66.5 & \textbf{67.7} \\
    \bottomrule
    
    \end{tabular}
    \caption{Quantization results of semantic segmentation on VOC dataset. We compress a Deeplab-V3 model with MobileNetv2 backbone and report the mean Intersection over Union (mIoU).}
    \label{deeplab}
\end{table}

\textbf{Results for Object Detection and Instance Segmentation.}
\begin{table}[h]
    \centering
    \captionsetup{skip=2pt}
    \renewcommand{\arraystretch}{0.8}
    \begin{minipage}{\linewidth}
    \centering
    \setlength{\tabcolsep}{1.5mm}
    \begin{tabular}{c|c|c|c|c}
    \toprule
    Model & C.R. & $AP_{50:95}^{bbox}$ & $AP_{50}^{bbox}$ & $AP_{75}^{bbox}$\\
    \midrule
    YOLOV9-S~\cite{wang2024yolov9}  & - & 46.8 & 63.4 & 41.3\\
    \midrule
    LSQ+BNR~\cite{esser2019learned} & 16$\times$ & 27.7$\dagger$ & 40.1$\dagger$ & 30.7$\dagger$ \\
    VQ  & 20$\times$ & 30.0 & 43.2 & 32.5\\
    SSVQ  & 20$\times$ & \textbf{35.2} & \textbf{49.7} & \textbf{38.1}\\
    \bottomrule
    \end{tabular}
    \end{minipage}

    \vspace{0.5em}
    
    \begin{minipage}{\linewidth}
    \centering
    \setlength{\tabcolsep}{1.0mm}
    \begin{tabular}{c|c|cc|cc}
    \toprule
    \multirow{2}{*}{Model} & \multirow{2}{*}{C.R.} & \multicolumn{2}{c|}{bbox AP} & \multicolumn{2}{c}{mask AP} \\
    & & $AP_{50:95}$ & $AP_{50}$ & $AP_{50:95}$ & $AP_{50}$\\
    \midrule
    GELAN-C~\cite{chang2023yolor}  & - & 52.3 & 69.2 & 42.4 & 65.9\\
    \midrule
    LSQ+BNR~\cite{esser2019learned} & 16 $\times$ & 18.1$\dagger$ & 39.5$\dagger$ & 15.8$\dagger$ & 28.9$\dagger$\\
    VQ & 21$\times$ & 43.5 & 59.9 & 34.6 & 55.9\\
    SSVQ & 21$\times$ & \textbf{47.6} & \textbf{64.4} & \textbf{38.5} & \textbf{60.8}\\
    \bottomrule
    \end{tabular}
    \end{minipage}
   \caption{Quantization results of object detection and instance segmentation on COCO dataset. 'bbox AP' is the bounding box average precision for object detection, and 'mask AP' is the mask average precision for instance segmentation. $\dagger$ denotes our reproduced results.}
   \label{detect}
   \vspace{-0.5cm}
\end{table}
Tab.~\ref{detect} presents object detection and instance segmentation results on COCO2017 dataset, where SSVQ achieves consistent 4$\sim$5\% mAP improvements over standard VQ at $\approx$20 $\times$ compression ratio, demonstrating robust performance across both tasks.

\begin{table}[h]
\renewcommand{\arraystretch}{0.7}
\setlength{\tabcolsep}{1.2mm}
\captionsetup{skip=2pt}
    \centering
    \begin{tabular}{ccccc}
    \toprule
Prompts & Method & Bit & FID-to-FP$\downarrow$  & CLIP$\uparrow$ \\
\cmidrule(lr){1-5}
\multicolumn{5}{c}{\textbf{Stable Diffusion v1-4}}\\
\cmidrule(lr){1-5}
\multirow{8}{*}{\makecell{COCO\\[-1pt]Prompts}} & FP & 16 & 0 & 31.4\\
& PCR~\cite{tang2024post} & 2 & 311.0$\dagger$ &  19.1 $\dagger$ \\ 
& EMF~\cite{li2024efficiency} & 2 & 173.4$\dagger$ & 18.3$\dagger$ \\

& VQ & 2 & 24.5 & 30.1  \\
& \textbf{SSVQ} & 2 & \textbf{13.6} & \textbf{30.8} \\
& Quest~\cite{wang2024quest}* & w2/a6 &  NA & 23.0*\\
& MPQDM\cite{feng2024mpq}* & w2/a6 &  NA & 25.1*\\
& \textbf{VQ*} & w2/a6 & 56.9 & 29.4* \\
& \textbf{SSVQ*} & w2/a6 & \textbf{31.3} & \textbf{30.6}* \\
\cmidrule(lr){2-5}
\multirow{5}{*}{\makecell{SD\\[-1pt]Prompts}} & FP & 16 & 0 & 31.6 \\
& PCR~\cite{tang2024post} & 2 & 322.6$\dagger$ & 19.4$\dagger$   \\
& EMF~\cite{li2024efficiency} & 2 & 102.8$\dagger$ & 22.3$\dagger$ \\
& VQ & 2 & 20.9 &  28.8 \\
& \textbf{SSVQ} & 2 & \textbf{15.0} & \textbf{30.2} \\
\cmidrule(lr){1-5}

\multicolumn{5}{c}{\textbf{Stable Diffusion v2-1}}\\
\cmidrule(lr){1-5}
\multirow{5}{*}{\makecell{COCO\\[-1pt]Prompts}} & FP & 16 & 0 & 31.5\\ 
& PCR~\cite{tang2024post} & 2 & 330.8$\dagger$ & 19.2$\dagger$  \\
& EMF~\cite{li2024efficiency} & 2 &137.6$\dagger$ & 23.6$\dagger$ \\

& VQ & 2 & 22.4 & 30.2 \\
& \textbf{SSVQ} & 2 & \textbf{15.0} & \textbf{31.1} \\
\cmidrule(lr){2-5}
\multirow{5}{*}{\makecell{SD\\[-1pt]Prompts}} & FP & 16 & 0 & 31.6\\ 
& PCR~\cite{tang2024post} & 2 & 364.6$\dagger$ & 20.1$\dagger$   \\
& EMF~\cite{li2024efficiency} & 2 & 181.7$\dagger$ & 21.3$\dagger$ \\

& VQ & 2 & 30.7 &  27.9 \\
& \textbf{SSVQ} & 2 & \textbf{19.1} & \textbf{29.6} \\
\midrule

\multicolumn{5}{c}{\textbf{Stable Diffusion 3}}\\
\midrule
\multirow{4}{*}{\makecell{COCO\\[-1pt]Prompts}} & FP & 16 & 0 & 31.7\\ & EMF~\cite{li2024efficiency} & 2 & 214.9$\dagger$ & 21.5$\dagger$\\
& VQ & 2 & 18.5 & 30.6 \\
& \textbf{SSVQ} & 2 & \textbf{11.2} & \textbf{31.7} \\
\cmidrule(lr){2-5}
\multirow{4}{*}{\makecell{SD\\[-1pt]Prompts}}  & FP & 16 & 0 & 30.6\\
& EMF~\cite{li2024efficiency} & 2 & 266.9$\dagger$ & 20.8$\dagger$\\
& VQ & 2 & 28.2 &  29.7\\
& \textbf{SSVQ} & 2 & \textbf{14.2} & \textbf{30.7} \\
\bottomrule
    \end{tabular}
    \caption{Quantization results of image generation of Stable Diffusion models on COCO 2017 val prompts and SD prompts. We generate 5000 images for each prompt set. CLIP scores were computed using ViT-B/16, with FP model baselines shown below prompts. $\dagger$ means reproduced results. $\downarrow$ means lower is better. $*$ means validation on COCO2014 val prompts.}
    \label{image generation}
\end{table}

\textbf{Results for Image Generation.} We evaluate SSVQ on two prompt datasets: COCO-2017 and Stable Diffusion prompts, testing both UNet-based (SDv1.4/SDv2.1) and transformer-based (SDv3) architectures. By default we only quantize weights, but to make a fair comparison with MPQDM~\cite{feng2024mpq}, we also provide w2a6 results. Under extreme 2-bit compression, where conventional scalar quantization fails, our vector quantization framework maintains functional generation capability. Quantitative evaluation using FID-to-FP~\cite{tang2024post} and CLIP score~\cite{hessel2021clipscore} shows SSVQ achieves 30\% lower FID-to-FP versus conventional VQ and 25\% higher CLIP score compared to SOTA mix-precision UQ approaches, demonstrating superior alignment with full-precision models. Additional visual results are included in the supplementary material.

\textbf{Results for NLP tasks.}
To further assess the generalization of our method, we validate it on Llama3.2-1B, using hessian-weighted k-means\cite{kim2023squeezellm} as initialization. Results in Tab.\ref{llm} clearly show SSVQ's superiority over VQ.
\begin{table}[t]
    \captionsetup{skip=2pt}
    \renewcommand{\arraystretch}{0.8}
    \setlength{\tabcolsep}{0.9mm}
    \centering
    \begin{tabular}{c|c|cc|ccc}
    \toprule
     Methods & bit/w & Wiki2$\downarrow$ & C4$\downarrow$ & WinoG$\uparrow$ & ArcE$\uparrow$ & ArcC$\uparrow$ \\
     \midrule
     FP &16 & 9.7 & 12.1 & 0.60 & 0.65 & 0.31\\
     \midrule
     VQ &  2.5 & 20.8 & 24.9 & 0.53 & 0.50 & 0.23\\
      SSVQ &  2.5 &  \textbf{11.9} &  \textbf{15.4} &  \textbf{0.57} &  \textbf{0.61} &  \textbf{0.29}\\
     \midrule
     VQ &  2 & 63.1 & 58.1 & 0.51 & 0.40 & 0.21\\
      SSVQ &  2 &  \textbf{13.9} &  \textbf{17.5} &  \textbf{0.54} &  \textbf{0.55} &  \textbf{0.27}\\
     \bottomrule
    \end{tabular}
    \caption{Comparison between VQ and SSVQ on Llama3.2-1B. We report PPL on Wiki2 \& C4, and accuracy for zero-shot tasks.}
    \label{llm}
    \vspace{-0.5cm}
\end{table}

\subsection{Ablation studies}
\begin{figure}[t]
    \centering
    \captionsetup{skip=2pt}
    \includegraphics[width=1\linewidth]{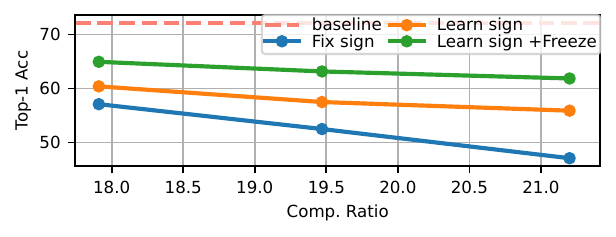}
    \caption{Ablation study on the effectiveness of the proposed learnable sign and improved freezing strategy, on the quantized DeiT-T model with k=16 and d=8}
    \label{ablation_ls}
\end{figure}

We conducted ablation experiments on the DeiT-tiny model. We first validate the impact of learnable signs and iterative freezing. As shown in Fig.~\ref{ablation_ls}, we demonstrated the accuracy-compression trade-off curves for three configurations: (1) fixed sign, (2) learnable signs without freezing, and (3) learnable signs with iterative freezing (Our SSVQ). The learnable sign bits provide consistent accuracy improvements ranging from 3\% to 9\%, with more significant gains observed under higher compression ratios. Meanwhile, the iterative freezing mechanism provides consistent 5\% accuracy improvements across compression rates. At a 21$\times$ compression ratio, SSVQ achieves an absolute 14.8\% accuracy gain over fixed-sign baselines, with progressive improvements evident in rate-distortion curves.

Next, we evaluate our improvement to iterative freezing strategy through Tab.~\ref{ablation_if}. Columns 1-4 demonstrate that optimized freezing frequencies improve accuracy. Signs can achieve a more precise state after experiencing brief fluctuations. However, excessively low freezing rates compromise training stability and final accuracy. Additionally, columns 3 and 5 reveal that majority sign voting better captures sign change trends compared to alternative approaches, providing more reliable frozen results.

\begin{table}[t]
    \centering
    \captionsetup{skip=2pt}
    \renewcommand{\arraystretch}{0.8}
    \setlength{\tabcolsep}{1.5mm}
    \begin{tabular}{c|c|c|c|c|c}
    \toprule
    Freeze iter & 100 & 200 & \textbf{500} & 1000 & 500\\
    \midrule
    Criterion & MSV & MSV & \textbf{MSV} & MSV & EMA\\
    \midrule
    Top-1 Acc & 60.9 & 61.47 & \textbf{61.90} &  61.55 & 60.8 \\
    \bottomrule
    \end{tabular}
    \caption{Ablation study on the selection of Freezing intervals and sign criterion. Experiments are conducted on the quantized DeiT-T model with k=16 and d=8. 'MSV' means majority sign voting, and 'EMA' means exponential moving average.}
    \label{ablation_if}
\end{table}

\section{System Evaluation}
\begin{figure}[h]
    \centering
    \captionsetup{skip=2pt}
    \includegraphics[width=0.95\linewidth]{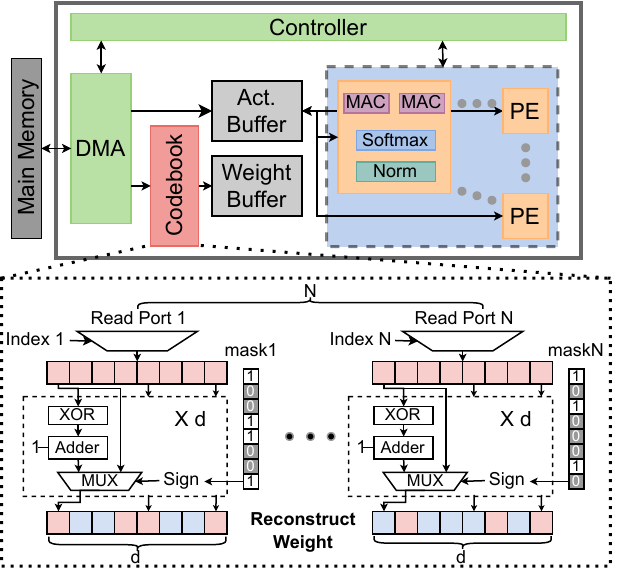}
    \caption{Architecture overview of our hardware simulator supporting SSVQ. }
    \label{arthitecture}
    \vspace{-0.2cm}
\end{figure}
\subsection{System setup}
We develop a hardware simulator by modifying AccelTran~\cite{tuli2023acceltran}, an originally NLP accelerator, through architectural adaptations for vision transformers. During implementation, we identified and resolved inconsistencies in main-memory data access within the open-source code.

The accelerator architecture is composed of three primary components: a controller module, a hierarchical memory system, and a parallel computation array. The memory subsystem features a three-tiered structure, incorporating off-chip main memory along with on-chip buffers for activation and weights, which enhances data reuse efficiency. At the core of the design is the computation array—an interconnected grid of processing engines (PEs). Each PE integrates multiply-accumulate (MAC) units to facilitate dense matrix operations, as well as dedicated hardware for nonlinear functions such as Softmax and LayerNorm. Detailed architectural specifications are provided in Tab.~\ref{design choice}.

\begin{table}[h]
    \centering
    \captionsetup{skip=2pt}
    \renewcommand{\arraystretch}{0.8}
    \begin{tabular}{c|c}
    \toprule
    Module & Configuration \\
    \midrule
    Main Memory & LP-DDR3, Bandwidth=128bit/cycle\\
    Buffer Size & Act: 0.5MB, Weight: 0.5MB\\
    PEs & 64 \\
    Compute Units & 16MAC Lanes/PE, 4Softmax unit/PE\\
    Codebook Size & k=256, d=8\\
    \bottomrule
    \end{tabular}
    \caption{Design choices of our hardware simulator.}
    \label{design choice}
    \vspace{-0.4cm}
\end{table}

\subsection{Architecture support for SSVQ}
To facilitate the inference of SSVQ models, we convert 1/-1 signs from the compression pipeline into binary 1/0 masks. Codebook elements are stored as 8-bit values (MSB=0, range 0-127). Codebook quantization brings negligible accuracy drops, as shown in the 'q\_ssvq' curve in DeiT-Tiny in Fig.~\ref{classify}.
To prevent misaligned memory access, we encode each index using 8 bits and utilize the [$\log_{2}(k)$:0] bits for codebook addressing.

The key design is our weight decoding module between the main memory and the weight buffer. The working flow is illustrated in the lower section of Fig.~\ref{arthitecture}. Firstly, layer-specific codebooks are loaded, followed by sign masks and assignments. Mask values determine weight reconstruction: positions with mask=0 are computed as two's complement negatives, while mask=1 preserves original values. The reconstructed weights are then stored in the weight buffer.

\subsection{Simulation results}
\begin{figure}[h]
    \centering
    \captionsetup{skip=2pt}
    \includegraphics[width=0.95\linewidth]{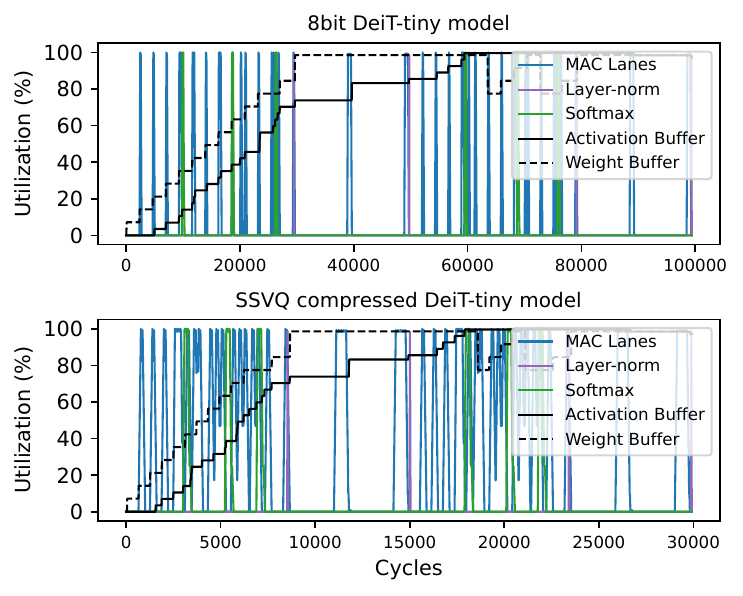}
    \caption{Hardware evaluation of SSVQ on the DeiT-Tiny. We report the resource utilization and running cycles.}
    \label{hardware_cycle}
    \vspace{-0.2cm}
\end{figure}

The DeiT-tiny model consists of 12 layers. Aside from the unique input embedding in the first layer, the behavior of the remaining layers is nearly identical. As a result, we will focus on the simulation outcomes for the first two layers. The findings for both the 8-bit compressed model and the SSVQ compressed model are illustrated in Fig.~\ref{hardware_cycle}.

The computational bottleneck related to weight loading is evident, particularly in the feed-forward network (FFN) layers. When compared to the 8-bit compressed model, the SSVQ model further reduces data access to the main memory, contributing to the observed 3× inference speed up. This demonstrates SSVQ's effectiveness in addressing hardware bottlenecks while maintaining model accuracy.
\section{Conclusion}
This paper presents Sign-Splitting Vector Quantization (SSVQ), a novel quantization framework that addresses key limitations in vector quantization during fine-tuning. Our approach introduces two core innovations: (1) learnable sign bits extracted from uncompressed weights that decouple weight and codeword update directions, and (2) an enhanced iterative freezing strategy that ensures stable training.

Extensive experiments across diverse vision tasks and architectures demonstrate SSVQ's superior accuracy-compression trade-offs compared to conventional VQ and UQ methods. Hardware simulations on DeiT-tiny validate SSVQ's practical efficiency, achieving 3 $\times$ inference acceleration. These results establish SSVQ as an effective solution for deploying compressed vision models on edge hardware.

\section{Acknowledgements}
    This work was supported by the National Natural Science Foundation of China (Grant No. 62274142), Sino-German Mobility Programme (Grant No. M-0499), and Zhejiang University - Vivo Information Technology Joint Research Center.

{
    \small
    \bibliographystyle{ieeenat_fullname}
    \bibliography{main}

\begin{thebibliography}{50}
\providecommand{\natexlab}[1]{#1}
\providecommand{\url}[1]{\texttt{#1}}
\expandafter\ifx\csname urlstyle\endcsname\relax
  \providecommand{\doi}[1]{doi: #1}\else
  \providecommand{\doi}{doi: \begingroup \urlstyle{rm}\Url}\fi

\bibitem[bubbliiiing(2023)]{bubbliiiing_deeplabv3_plus_pytorch}
bubbliiiing.
\newblock Deeplabv3+ in pytorch.
\newblock \url{https://github.com/bubbliiiing/deeplabv3-plus-pytorch}, 2023.
\newblock Accessed: October 10, 2023.

\bibitem[Chang et~al.(2023)Chang, Wang, Wang, Chou, and Liao]{chang2023yolor}
Hung-Shuo Chang, Chien-Yao Wang, Richard~Robert Wang, Gene Chou, and Hong-Yuan~Mark Liao.
\newblock Yolor-based multi-task learning.
\newblock \emph{arXiv preprint arXiv:2309.16921}, 2023.

\bibitem[Chen(2017)]{chen2017rethinking}
Liang-Chieh Chen.
\newblock Rethinking atrous convolution for semantic image segmentation.
\newblock \emph{arXiv preprint arXiv:1706.05587}, 2017.

\bibitem[Chen et~al.(2020)Chen, Wang, and Cheng]{chen2020towards}
Weihan Chen, Peisong Wang, and Jian Cheng.
\newblock Towards convolutional neural networks compression via global\&progressive product quantization.
\newblock In \emph{BMVC}, 2020.

\bibitem[Chen et~al.(2022)Chen, Dai, Chen, Liu, Dong, Yuan, and Liu]{chen2022mobile}
Yinpeng Chen, Xiyang Dai, Dongdong Chen, Mengchen Liu, Xiaoyi Dong, Lu Yuan, and Zicheng Liu.
\newblock Mobile-former: Bridging mobilenet and transformer.
\newblock In \emph{Proceedings of the IEEE/CVF conference on computer vision and pattern recognition}, pages 5270--5279, 2022.

\bibitem[Cho et~al.(2022)Cho, Vahid, Adya, and Rastegari]{cho2022dkm}
Minsik Cho, Keivan~A. Vahid, Saurabh Adya, and Mohammad Rastegari.
\newblock Dkm: Differentiable k-means clustering layer for neural network compression, 2022.

\bibitem[Esser et~al.(2019)Esser, McKinstry, Bablani, Appuswamy, and Modha]{esser2019learned}
Steven~K Esser, Jeffrey~L McKinstry, Deepika Bablani, Rathinakumar Appuswamy, and Dharmendra~S Modha.
\newblock Learned step size quantization.
\newblock \emph{arXiv preprint arXiv:1902.08153}, 2019.

\bibitem[Feng et~al.(2024)Feng, Qin, Yang, An, Huang, Diao, Wang, Tao, Xu, and Magno]{feng2024mpq}
Weilun Feng, Haotong Qin, Chuanguang Yang, Zhulin An, Libo Huang, Boyu Diao, Fei Wang, Renshuai Tao, Yongjun Xu, and Michele Magno.
\newblock Mpq-dm: Mixed precision quantization for extremely low bit diffusion models.
\newblock \emph{arXiv preprint arXiv:2412.11549}, 2024.

\bibitem[Gong et~al.(2014)Gong, Liu, Yang, and Bourdev]{gong2014compressing}
Yunchao Gong, Liu Liu, Ming Yang, and Lubomir Bourdev.
\newblock Compressing deep convolutional networks using vector quantization.
\newblock \emph{arXiv preprint arXiv:1412.6115}, 2014.

\bibitem[Han et~al.(2015)Han, Mao, and Dally]{han2015deep}
Song Han, Huizi Mao, and William~J Dally.
\newblock Deep compression: Compressing deep neural networks with pruning, trained quantization and huffman coding.
\newblock \emph{arXiv preprint arXiv:1510.00149}, 2015.

\bibitem[Hessel et~al.(2021)Hessel, Holtzman, Forbes, Bras, and Choi]{hessel2021clipscore}
Jack Hessel, Ari Holtzman, Maxwell Forbes, Ronan~Le Bras, and Yejin Choi.
\newblock Clipscore: A reference-free evaluation metric for image captioning.
\newblock \emph{arXiv preprint arXiv:2104.08718}, 2021.

\bibitem[Howard et~al.(2019)Howard, Sandler, Chu, Chen, Chen, Tan, Wang, Zhu, Pang, Vasudevan, et~al.]{howard2019searching}
Andrew Howard, Mark Sandler, Grace Chu, Liang-Chieh Chen, Bo Chen, Mingxing Tan, Weijun Wang, Yukun Zhu, Ruoming Pang, Vijay Vasudevan, et~al.
\newblock Searching for mobilenetv3.
\newblock In \emph{Proceedings of the IEEE/CVF international conference on computer vision}, pages 1314--1324, 2019.

\bibitem[Hu et~al.(2025)Hu, Wang, Chen, Zhang, Wang, Li, and Nan]{hu2025dynamicid}
Xirui Hu, Jiahao Wang, Hao Chen, Weizhan Zhang, Benqi Wang, Yikun Li, and Haishun Nan.
\newblock Dynamicid: Zero-shot multi-id image personalization with flexible facial editability.
\newblock \emph{arXiv preprint arXiv:2503.06505}, 2025.

\bibitem[Jegou et~al.(2010)Jegou, Douze, and Schmid]{jegou2010product}
Herve Jegou, Matthijs Douze, and Cordelia Schmid.
\newblock Product quantization for nearest neighbor search.
\newblock \emph{IEEE transactions on pattern analysis and machine intelligence}, 33\penalty0 (1):\penalty0 117--128, 2010.

\bibitem[Kim et~al.(2023)Kim, Hooper, Gholami, Dong, Li, Shen, Mahoney, and Keutzer]{kim2023squeezellm}
Sehoon Kim, Coleman Hooper, Amir Gholami, Zhen Dong, Xiuyu Li, Sheng Shen, Michael~W Mahoney, and Kurt Keutzer.
\newblock Squeezellm: Dense-and-sparse quantization.
\newblock \emph{arXiv preprint arXiv:2306.07629}, 2023.

\bibitem[Kuzmin et~al.(2023)Kuzmin, Nagel, van Baalen, Behboodi, and Blankevoort]{kuzmin2023pruning}
Andrey Kuzmin, Markus Nagel, Mart van Baalen, Arash Behboodi, and Tijmen Blankevoort.
\newblock Pruning vs quantization: Which is better?, 2023.

\bibitem[Li et~al.(2016)Li, Kadav, Durdanovic, Samet, and Graf]{li2016pruning}
Hao Li, Asim Kadav, Igor Durdanovic, Hanan Samet, and Hans~Peter Graf.
\newblock Pruning filters for efficient convnets.
\newblock \emph{arXiv preprint arXiv:1608.08710}, 2016.

\bibitem[Li et~al.(2024)Li, Deng, Wang, Gu, Xu, Shen, and Huang]{li2024efficiency}
Shuaiting Li, Juncan Deng, Zeyu Wang, Hong Gu, Kedong Xu, Haibin Shen, and Kejie Huang.
\newblock Efficiency meets fidelity: A novel quantization framework for stable diffusion.
\newblock \emph{arXiv preprint arXiv:2412.06661}, 2024.

\bibitem[Li et~al.(2025)Li, Wang, Deng, Wang, Ye, Wang, Shen, and Huang]{li2025mvq}
Shuaiting Li, Chengxuan Wang, Juncan Deng, Zeyu Wang, Zewen Ye, Zongsheng Wang, Haibin Shen, and Kejie Huang.
\newblock Mvq: Towards efficient dnn compression and acceleration with masked vector quantization.
\newblock In \emph{Proceedings of the 30th ACM International Conference on Architectural Support for Programming Languages and Operating Systems, Volume 1}, pages 731--745, 2025.

\bibitem[Liu et~al.(2023{\natexlab{a}})Liu, Chen, Ma, and Liu]{liu2023hyperspherical}
Dan Liu, Xi Chen, Chen Ma, and Xue Liu.
\newblock Hyperspherical quantization: Toward smaller and more accurate models.
\newblock In \emph{Proceedings of the IEEE/CVF Winter Conference on Applications of Computer Vision}, pages 5262--5272, 2023{\natexlab{a}}.

\bibitem[Liu et~al.(2023{\natexlab{b}})Liu, Liu, and Cheng]{liu2023oscillation}
Shih-Yang Liu, Zechun Liu, and Kwang-Ting Cheng.
\newblock Oscillation-free quantization for low-bit vision transformers.
\newblock In \emph{International conference on machine learning}, pages 21813--21824. PMLR, 2023{\natexlab{b}}.

\bibitem[Liu et~al.(2023{\natexlab{c}})Liu, Peng, Zheng, Yang, Hu, and Yuan]{liu2023efficientvit}
Xinyu Liu, Houwen Peng, Ningxin Zheng, Yuqing Yang, Han Hu, and Yixuan Yuan.
\newblock Efficientvit: Memory efficient vision transformer with cascaded group attention.
\newblock In \emph{Proceedings of the IEEE/CVF Conference on Computer Vision and Pattern Recognition}, pages 14420--14430, 2023{\natexlab{c}}.

\bibitem[Liu et~al.(2018)Liu, Wu, Luo, Yang, Liu, and Cheng]{liu2018bi}
Zechun Liu, Baoyuan Wu, Wenhan Luo, Xin Yang, Wei Liu, and Kwang-Ting Cheng.
\newblock Bi-real net: Enhancing the performance of 1-bit cnns with improved representational capability and advanced training algorithm.
\newblock In \emph{Proceedings of the European conference on computer vision (ECCV)}, pages 722--737, 2018.

\bibitem[Liu et~al.(2022)Liu, Mao, Wu, Feichtenhofer, Darrell, and Xie]{liu2022convnet}
Zhuang Liu, Hanzi Mao, Chao-Yuan Wu, Christoph Feichtenhofer, Trevor Darrell, and Saining Xie.
\newblock A convnet for the 2020s.
\newblock In \emph{Proceedings of the IEEE/CVF conference on computer vision and pattern recognition}, pages 11976--11986, 2022.

\bibitem[Lloyd(1982)]{lloyd1982least}
Stuart Lloyd.
\newblock Least squares quantization in pcm.
\newblock \emph{IEEE transactions on information theory}, 28\penalty0 (2):\penalty0 129--137, 1982.

\bibitem[Louizos et~al.(2017)Louizos, Welling, and Kingma]{louizos2017learning}
Christos Louizos, Max Welling, and Diederik~P Kingma.
\newblock Learning sparse neural networks through $ l\_0 $ regularization.
\newblock \emph{arXiv preprint arXiv:1712.01312}, 2017.

\bibitem[MacQueen(1967)]{macqueen1967some}
James MacQueen.
\newblock Some methods for classification and analysis of multivariate observations.
\newblock In \emph{Proceedings of the Fifth Berkeley Symposium on Mathematical Statistics and Probability, Volume 1: Statistics}, pages 281--298. University of California press, 1967.

\bibitem[Martinez et~al.(2021)Martinez, Shewakramani, Liu, Barsan, Zeng, and Urtasun]{Martinez_2021_CVPR}
Julieta Martinez, Jashan Shewakramani, Ting~Wei Liu, Ioan~Andrei Barsan, Wenyuan Zeng, and Raquel Urtasun.
\newblock Permute, quantize, and fine-tune: Efficient compression of neural networks.
\newblock In \emph{Proceedings of the IEEE/CVF Conference on Computer Vision and Pattern Recognition (CVPR)}, pages 15699--15708, 2021.

\bibitem[Mehta and Rastegari(2021)]{mehta2021mobilevit}
Sachin Mehta and Mohammad Rastegari.
\newblock Mobilevit: light-weight, general-purpose, and mobile-friendly vision transformer.
\newblock \emph{arXiv preprint arXiv:2110.02178}, 2021.

\bibitem[Molchanov et~al.(2019)Molchanov, Mallya, Tyree, Frosio, and Kautz]{molchanov2019importance}
Pavlo Molchanov, Arun Mallya, Stephen Tyree, Iuri Frosio, and Jan Kautz.
\newblock Importance estimation for neural network pruning.
\newblock In \emph{Proceedings of the IEEE/CVF conference on computer vision and pattern recognition}, pages 11264--11272, 2019.

\bibitem[Nagel et~al.(2022)Nagel, Fournarakis, Bondarenko, and Blankevoort]{nagel2022overcoming}
Markus Nagel, Marios Fournarakis, Yelysei Bondarenko, and Tijmen Blankevoort.
\newblock Overcoming oscillations in quantization-aware training.
\newblock In \emph{International Conference on Machine Learning}, pages 16318--16330. PMLR, 2022.

\bibitem[Pullini et~al.(2019)Pullini, Rossi, Loi, Tagliavini, and Benini]{pullini2019mr}
Antonio Pullini, Davide Rossi, Igor Loi, Giuseppe Tagliavini, and Luca Benini.
\newblock Mr. wolf: An energy-precision scalable parallel ultra low power soc for iot edge processing.
\newblock \emph{IEEE Journal of Solid-State Circuits}, 54\penalty0 (7):\penalty0 1970--1981, 2019.

\bibitem[Qin et~al.(2024)Qin, Leichner, Delakis, Fornoni, Luo, Yang, Wang, Banbury, Ye, Akin, et~al.]{qin2024mobilenetv4}
Danfeng Qin, Chas Leichner, Manolis Delakis, Marco Fornoni, Shixin Luo, Fan Yang, Weijun Wang, Colby Banbury, Chengxi Ye, Berkin Akin, et~al.
\newblock Mobilenetv4: Universal models for the mobile ecosystem.
\newblock In \emph{European Conference on Computer Vision}, pages 78--96. Springer, 2024.

\bibitem[Renda et~al.(2020)Renda, Frankle, and Carbin]{renda2020comparing}
Alex Renda, Jonathan Frankle, and Michael Carbin.
\newblock Comparing rewinding and fine-tuning in neural network pruning.
\newblock \emph{arXiv preprint arXiv:2003.02389}, 2020.

\bibitem[Rombach et~al.(2022)Rombach, Blattmann, Lorenz, Esser, and Ommer]{rombach2022high}
Robin Rombach, Andreas Blattmann, Dominik Lorenz, Patrick Esser, and Bj{\"o}rn Ommer.
\newblock High-resolution image synthesis with latent diffusion models.
\newblock In \emph{Proceedings of the IEEE/CVF conference on computer vision and pattern recognition}, pages 10684--10695, 2022.

\bibitem[Rossi et~al.(2021)Rossi, Conti, Eggiman, Mach, Di~Mauro, Guermandi, Tagliavini, Pullini, Loi, Chen, et~al.]{rossi20214}
Davide Rossi, Francesco Conti, Manuel Eggiman, Stefan Mach, Alfio Di~Mauro, Marco Guermandi, Giuseppe Tagliavini, Antonio Pullini, Igor Loi, Jie Chen, et~al.
\newblock 4.4 a 1.3 tops/w@ 32gops fully integrated 10-core soc for iot end-nodes with 1.7 $\mu$w cognitive wake-up from mram-based state-retentive sleep mode.
\newblock In \emph{2021 IEEE International Solid-State Circuits Conference (ISSCC)}, pages 60--62. IEEE, 2021.

\bibitem[Sandler et~al.(2018)Sandler, Howard, Zhu, Zhmoginov, and Chen]{sandler2018mobilenetv2}
Mark Sandler, Andrew Howard, Menglong Zhu, Andrey Zhmoginov, and Liang-Chieh Chen.
\newblock Mobilenetv2: Inverted residuals and linear bottlenecks.
\newblock In \emph{Proceedings of the IEEE conference on computer vision and pattern recognition}, pages 4510--4520, 2018.

\bibitem[Son et~al.(2018)Son, Nah, and Lee]{son2018clustering}
Sanghyun Son, Seungjun Nah, and Kyoung~Mu Lee.
\newblock Clustering convolutional kernels to compress deep neural networks.
\newblock In \emph{Proceedings of the European conference on computer vision (ECCV)}, pages 216--232, 2018.

\bibitem[Stock et~al.(2019)Stock, Joulin, Gribonval, Graham, and J{\'e}gou]{stock2019and}
Pierre Stock, Armand Joulin, R{\'e}mi Gribonval, Benjamin Graham, and Herv{\'e} J{\'e}gou.
\newblock And the bit goes down: Revisiting the quantization of neural networks.
\newblock \emph{arXiv preprint arXiv:1907.05686}, 2019.

\bibitem[Tan and Le(2019)]{tan2019efficientnet}
Mingxing Tan and Quoc Le.
\newblock Efficientnet: Rethinking model scaling for convolutional neural networks.
\newblock In \emph{International conference on machine learning}, pages 6105--6114. PMLR, 2019.

\bibitem[Tang et~al.(2024)Tang, Wang, Chen, Guan, Wu, Tang, and Zhu]{tang2024post}
Siao Tang, Xin Wang, Hong Chen, Chaoyu Guan, Zewen Wu, Yansong Tang, and Wenwu Zhu.
\newblock Post-training quantization with progressive calibration and activation relaxing for text-to-image diffusion models.
\newblock In \emph{European Conference on Computer Vision}, pages 404--420. Springer, 2024.

\bibitem[Touvron et~al.(2021)Touvron, Cord, Douze, Massa, Sablayrolles, and J{\'e}gou]{touvron2021training}
Hugo Touvron, Matthieu Cord, Matthijs Douze, Francisco Massa, Alexandre Sablayrolles, and Herv{\'e} J{\'e}gou.
\newblock Training data-efficient image transformers \& distillation through attention.
\newblock In \emph{International conference on machine learning}, pages 10347--10357. PMLR, 2021.

\bibitem[Tuli and Jha(2023)]{tuli2023acceltran}
Shikhar Tuli and Niraj~K Jha.
\newblock Acceltran: A sparsity-aware accelerator for dynamic inference with transformers.
\newblock \emph{IEEE Transactions on Computer-Aided Design of Integrated Circuits and Systems}, 42\penalty0 (11):\penalty0 4038--4051, 2023.

\bibitem[van Baalen et~al.(2024)van Baalen, Kuzmin, Nagel, Couperus, Bastoul, Mahurin, Blankevoort, and Whatmough]{van2024gptvq}
Mart van Baalen, Andrey Kuzmin, Markus Nagel, Peter Couperus, Cedric Bastoul, Eric Mahurin, Tijmen Blankevoort, and Paul Whatmough.
\newblock Gptvq: The blessing of dimensionality for llm quantization.
\newblock \emph{arXiv preprint arXiv:2402.15319}, 2024.

\bibitem[Vasu et~al.(2023)Vasu, Gabriel, Zhu, Tuzel, and Ranjan]{vasu2023fastvit}
Pavan Kumar~Anasosalu Vasu, James Gabriel, Jeff Zhu, Oncel Tuzel, and Anurag Ranjan.
\newblock Fastvit: A fast hybrid vision transformer using structural reparameterization.
\newblock In \emph{Proceedings of the IEEE/CVF International Conference on Computer Vision}, pages 5785--5795, 2023.

\bibitem[Wang et~al.(2024{\natexlab{a}})Wang, Yeh, and Mark~Liao]{wang2024yolov9}
Chien-Yao Wang, I-Hau Yeh, and Hong-Yuan Mark~Liao.
\newblock Yolov9: Learning what you want to learn using programmable gradient information.
\newblock In \emph{European conference on computer vision}, pages 1--21. Springer, 2024{\natexlab{a}}.

\bibitem[Wang et~al.(2024{\natexlab{b}})Wang, Shang, Yuan, Wu, Yan, and Yan]{wang2024quest}
Haoxuan Wang, Yuzhang Shang, Zhihang Yuan, Junyi Wu, Junchi Yan, and Yan Yan.
\newblock Quest: Low-bit diffusion model quantization via efficient selective finetuning.
\newblock \emph{arXiv preprint arXiv:2402.03666}, 2024{\natexlab{b}}.

\bibitem[Wu et~al.(2016)Wu, Leng, Wang, Hu, and Cheng]{wu2016quantized}
Jiaxiang Wu, Cong Leng, Yuhang Wang, Qinghao Hu, and Jian Cheng.
\newblock Quantized convolutional neural networks for mobile devices.
\newblock In \emph{Proceedings of the IEEE conference on computer vision and pattern recognition}, pages 4820--4828, 2016.

\bibitem[Yu et~al.(2024)Yu, Zhou, Yan, and Wang]{yu2024inceptionnext}
Weihao Yu, Pan Zhou, Shuicheng Yan, and Xinchao Wang.
\newblock Inceptionnext: When inception meets convnext.
\newblock In \emph{Proceedings of the IEEE/CVF Conference on Computer Vision and Pattern Recognition}, pages 5672--5683, 2024.

\bibitem[Zhu et~al.(2023)Zhu, Dong, and Zhao]{zhu2023learning}
Zezhou Zhu, Yuan Dong, and Zhong Zhao.
\newblock Learning low-rank representations for model compression.
\newblock In \emph{2023 International Joint Conference on Neural Networks (IJCNN)}, pages 1--9. IEEE, 2023.

\end{thebibliography}
}

\clearpage
\setcounter{page}{1}
\maketitlesupplementary

\section{Analysis of initial clustering error}
\label{sec:initial error}
\begin{figure}[t]
    \centering
    \includegraphics[width=1\linewidth]{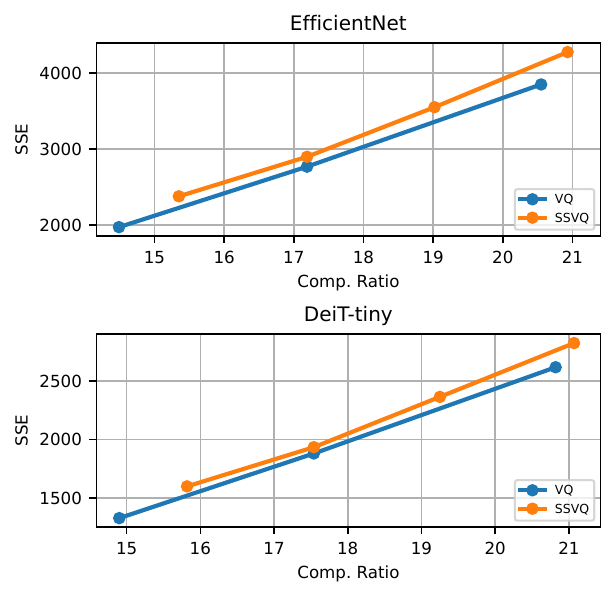}
    \caption{Comparison of initial clustering error between VQ and SSVQ.}
    \label{mse}
\end{figure}
Although storing the sign bit requires an additional 1 bit, our method does not introduce extra storage overhead. After extracting the sign bits, we can cluster the absolute values of the weights. Naturally, the similarity among non-negative weights is higher, so we only need fewer bits for cluster indexing. In Fig.~\ref{mse}, we compare the initial clustering errors of SSVQ and VQ. The results show that under the same compression rate, their clustering errors are comparable, which validates our hypothesis.

\section{Further analysis with dimension}
We further investigate the impact of quantization dimensionality. While prior work~\cite{stock2019and, Martinez_2021_CVPR, van2024gptvq, li2025mvq} has demonstrated that increasing quantization dimensionality improves model accuracy under fixed index bits, this approach introduces significant scalability challenges. Specifically, when doubling the dimensionality from $d$ to $2\times d$ with constant index bits, the codebook size expands exponentially by $2^d$ times. This substantial expansion makes the codebook a critical factor in compression ratio computation, potentially offsetting the benefits of higher-dimensional quantization.

The codebook storage impact is particularly pronounced in compact architectures like MobileNet and DeiT-Tiny. As shown in Fig.~\ref{classify}, VQ achieves optimal performance at d=4, while SSVQ performs best at d=8. For larger models (DeiT-Small and ConvNext-Tiny), VQ demonstrates better accuracy-compression trade-offs at d=8. To further explore SSVQ's scalability, we present additional results at d=16 for these larger architectures, revealing its superior adaptability to higher-dimensional quantization.

\section{Effect of Hessian weighted k-means}
SqueezeLLM~\cite{kim2023squeezellm} employs Hessian-weighted initialization. Specifically, its objective function is designed to minimize the final task loss caused by quantized weights
\begin{equation}
    \delta L = L(W) - L(\hat{W})
    \label{loss}
\end{equation}

We can achieve this objective through Hessian matrix approximation. Assuming that $\hat{W}$ represents a small perturbation on W, we perform a second-order Taylor expansion on Eq.~\ref{loss} to obtain:
\begin{equation}
    \delta L \approx g^T(W-\hat{W}) + \frac{1}{2}(W-\hat{W})^{T}H(W-\hat{W})
    \label{hessian}
\end{equation}

Here, $g$ is the gradient, and $H$ is the Hessian matrix of the weights (the second derivative of the Loss with respect to the weights). Assuming the pretrained model has converged, the gradient can be approximated as zero.
Furthermore, we can approximate $H$ by retaining only its diagonal elements (which typically dominate).
\begin{equation}
    \delta L \approx \sum_{i} H_{ii} \| \mathbf{w}_i - \mathbf{c}_{A(i)} \|^2
    \label{weighted}
\end{equation}

We observed that this approach significantly benefits LLMs, but fails to demonstrate clear advantages over standard k-means on vision-related tasks, especially after fine-tuning. Thus, we retain standard k-means in other experiments.

\section{Detailed quantization settings.}
\label{sec:quant settings}
In this section, we listed the detailed quantization settings of our experiments, specifically $k$ and $d$. Classification tasks (Tab.~\ref{table: different compression regimes}), detection and segmentation tasks (Tab.~\ref{tab:settings_detection}), image generation and nlp tasks (Tab.~\ref{tab:settings_gen_nlp}).
\begin{table}[t]
    \centering
    \begin{tabular}{c|c|c|c|c|c}
    \toprule
    \multicolumn{6}{c}{\textbf{SSVQ, d=8}} \\
    \midrule
    k & 8 & 16 & 32 & 64 & 128\\
    \midrule
    MobileNet-v2 &22.9 & 20.7 & 18.6 & 16.6 & - \\
    MobileNet-v3 & 23.2 & 21.3 & 19.6 & 18.1 & 16.7\\
    EfficientNet-lite0 & 23.0 & 20.9 & 19.0 & 17.2 & -\\
    ConvNext-Tiny & 23.2 & 21.3 & 19.6 & 18.1 & 16.8 \\
    DeiT-Tiny & 23.1 & 21.1 & 19.2 & 17.5& 15.8\\
    DeiT-Small & 23.2 & 21.3 & 19.6 & 18.1 & - \\
    \midrule
    
    \multicolumn{6}{c}{\textbf{VQ, d=4}} \\
    \midrule
    k & 32 & 64 & 128 & 256 & -\\
    \midrule
    MobileNet-v2 & - & 20.1 & 16.6 & - & - \\
    MobileNet-v3 & 23.5 & 18.7 & - & -& - \\
    EfficientNet-lite0 & - & 20.6 & 17.2 & - & - \\
    ConvNext-Tiny & - & 21.2 & 18.1 & - & -\\
    DeiT-Tiny & -& 20.8 & 17.5 & - & - \\
    DeiT-Small & - & 21.2 & 18.1 & - & - \\
    \midrule

    \multicolumn{6}{c}{\textbf{VQ, d=8}} \\
    \midrule
    k & 128 & 256 & 512 & 1024 & 2048 \\
    \midrule
    MobileNet-v2 & -& -& 20.9& 16.2& - \\
    MobileNet-v3 & 19.8 & - & -& -& - \\
    EfficientNet-lite0 & - & & 17.8 & - & -\\
    ConvNext-Tiny & -& -& 25.6& 21.6& 18.1\\
    DeiT-Tiny & - & 24.7 & 18.6 & - & - \\
    DeiT-Small & - & -& -& 25.1& 20.7\\

    \bottomrule
    \end{tabular}
    \caption{$k$\&$d$ settings and corresponding compression ratios of classification tasks (Fig.~\ref{classify}). Notably, results with compression ratios $\leq$16× or unacceptably low accuracy are omitted from the figure.}
    \label{table: different compression regimes}
\end{table}

\begin{table}[t]
    \centering
    \begin{tabular}{c|c|c|c|c}
    \toprule
    Models & Methods & k & d & C.R.\\
    \midrule
    \multirow{2}{*}{Yolo-v9} & VQ & 64 & 4 & 19.67\\
    & SSVQ & 16 & 8 & 20.47\\
    \midrule
    \multirow{2}{*}{Gelan-C} & VQ & 64 & 4 & 20.98\\
    & SSVQ & 16 & 8 & 21.15\\
    \end{tabular}
    \caption{$k$\&$d$ settings and corresponding bits/weight of detection and segmentation tasks (Tab.~\ref{detect}).}
    \label{tab:settings_detection}
\end{table}

\begin{table}[t]
    \centering
    \setlength{\tabcolsep}{1.5mm}
    \begin{tabular}{c|c|c|c|c|c}
    \toprule
    Methods & k & d & Index bit & Mask bit & Total bit\\
    \midrule
    VQ & 1024 & 4 & 2.5 & 0 & 2.5\\
    VQ & 256 & 4 & 2 & 0 & 2\\
    SSVQ & 64 & 4 & 1.5 & 1 & 2.5\\
    SSVQ & 256 & 8 & 1 & 1 & 2\\
    \end{tabular}
    \caption{$k$ and $d$ settings and corresponding bits/weight of generation (Tab.~\ref{image generation}) and NLP tasks (Tab.~\ref{llm}).}
    \label{tab:settings_gen_nlp}
\end{table}

\section{More results on image generation tasks}
\subsection{Comparison on visual similarity}
\begin{table}[h]
    \centering
    \begin{tabular}{ccccc}
    \toprule
    Prompts & Methods & LPIPS$\downarrow$ & SSIM$\uparrow$ & PSNR $\uparrow$ \\
    \midrule
    \multicolumn{5}{c}{\textbf{Stable Diffusion v1-4}}\\
    \midrule
    \multirow{3}{*}{\makecell{COCO\\Prompts}} & EMF~\cite{li2024efficiency} & 0.77 & 0.34& 11.1\\
    & VQ & 0.56 & 0.44 & 12.5\\
    & SSVQ & 0.51 & 0.49 & 13.4 \\
    \cmidrule{2-5}
    \multirow{3}{*}{\makecell{SD\\Prompts}} & EMF~\cite{li2024efficiency} & 0.66 & 0.42& 12.8\\
    & VQ & 0.56 & 0.46 & 13.5\\
    & SSVQ & 0.49 & 0.51 & 14.8 \\
    \midrule
    \multicolumn{5}{c}{\textbf{Stable Diffusion v2-1}}\\
    \midrule
    \multirow{3}{*}{\makecell{COCO\\Prompts}} & EMF~\cite{li2024efficiency} & 0.66 & 0.39 & 11.5\\
    & VQ & 0.59 & 0.44 & 12.7\\
    & SSVQ & 0.48 & 0.50 & 13.8 \\
    \cmidrule{2-5}
    \multirow{3}{*}{\makecell{SD\\Prompts}} & EMF~\cite{li2024efficiency} & 0.72 & 0.30 & 10.9\\
    & VQ & 0.64 & 0.46 & 13.3\\
    & SSVQ & 0.52 & 0.48 & 14.4 \\
    \midrule
    \multicolumn{5}{c}{\textbf{Stable Diffusion v3}}\\
    \midrule
    \multirow{3}{*}{\makecell{COCO\\Prompts}} &
    EMF~\cite{li2024efficiency} & 0.72 & 0.38 & 9.6\\
    & VQ & 0.59 & 0.42 & 10.4\\
    & SSVQ & 0.54 & 0.48 & 11.0 \\
    \cmidrule{2-5}
    \multirow{3}{*}{\makecell{SD\\Prompts}} &
    EMF~\cite{li2024efficiency} & 0.75 & 0.28 & 6.80\\
    & VQ & 0.64 & 0.34 & 9.40\\
    & SSVQ & 0.58 & 0.40 & 9.90 \\
    \bottomrule
    \end{tabular}
    \caption{Visual similarity results on three stable diffusion models. $\downarrow$ means lower is better. $\uparrow$ means higher is better}
    \label{visual_similarity}
\end{table}
To provide a comprehensive evaluation of generation consistency, we introduce visual similarity metrics as complementary assessment criteria. Specifically, we employ three widely-adopted measures: SSIM (structural similarity), LPIPS (learned perceptual image patch similarity), and PSNR (peak signal-to-noise ratio), maintaining consistency with the evaluation protocol used in EMF~\cite{li2024efficiency}. As demonstrated in Tab.~\ref{visual_similarity}, our SSVQ framework achieves an average improvement of 10\% in visual similarity metrics compared to the standard VQ baseline, highlighting its enhanced capability in preserving visual fidelity.
\subsection{Visual results}
\begin{figure*}
    \centering
    \includegraphics[width=0.9\linewidth]{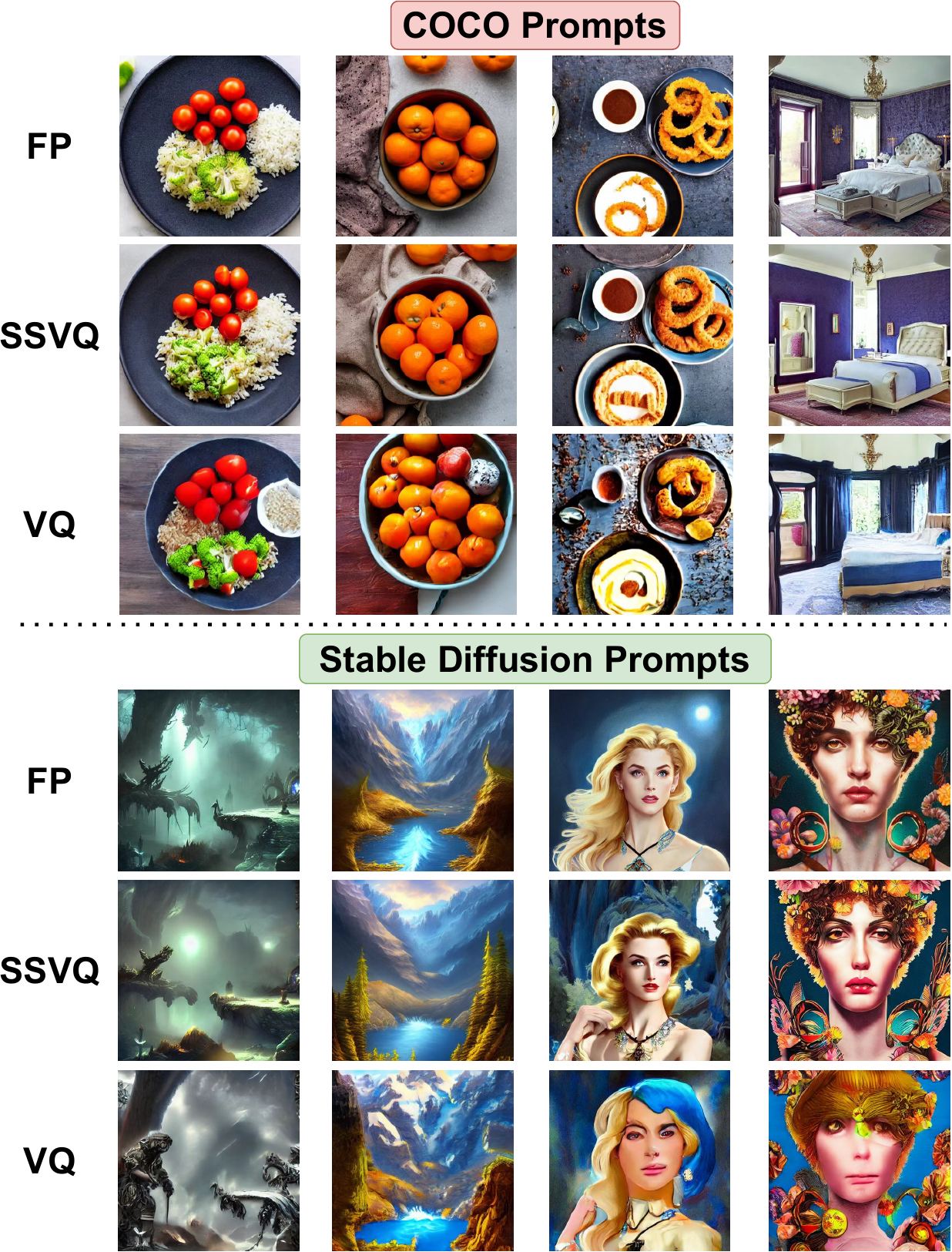}
    \caption{Visualization of image generation results of full-precision and quantized Stable Diffusion models.  Qualitative comparisons under COCO and Stable Diffusion prompt styles demonstrate SSVQ's superior generation quality and prompt consistency over conventional VQ.}
    \label{sd14_ssvq}
\end{figure*}

To provide intuitive visual comparisons, Fig.~\ref{sd14_ssvq} illustrates the qualitative results under extreme low-bit quantization. Our analysis reveals that aggressive quantization in diffusion models leads to three characteristic degradation patterns: (1) structural distortion in scene layout, (2) color bleeding and object disappearance, and (3) object blending artifacts. Notably, our SSVQ framework demonstrates remarkable robustness against these degradation effects, significantly mitigating image quality deterioration even at ultra-low bit rates.

\begin{table*}[t]
    \centering
    \begin{tabular}{c|c|c|c|c|c|c}
    \toprule
    Models & batch size & lr & lr\_mask & epochs & $\alpha$ & Optimizer\\
    \midrule
    \multicolumn{7}{c}{Classification}\\
    \midrule
    MobileNet-V2 & 256 & 2e-3 & 2e-4 & 10 & 1 & AdamW\\
    MobileNet-V3 & 256 & 2e-3 & 2e-4 & 10 & 1& AdamW\\
    EfficientNet-Lite0 & 256 & 2e-3 & 2e-4 & 10 & 1& AdamW\\
    ConvNext-Tiny & 256 & 2.5e-4 & 2.5e-4 & 10 & 4 & AdamW\\
    DeiT-Tiny & 256 & 2.5e-4 & 2.5e-4 & 10 & 6 & AdamW\\
    DeiT-Small & 256 & 2.5e-4 & 2.5e-4 & 10 & 4 & AdamW\\
    \midrule
    \multicolumn{7}{c}{Semantic Segmantation} \\
    \midrule
    DeepLab-V3 & 8 & 5e-4 & 5e-4& 100 & 10& AdamW \\
    \midrule
    \multicolumn{7}{c}{Object Dection} \\
    \midrule
    YOLO-V9 & 16 & 1e-3 & 1e-3& 10& 4 & AdamW\\
    \midrule
    \multicolumn{7}{c}{Instance Segmantation} \\
    \midrule
    GELAN-C-SEG & 32 & 1e-3 & 1e-4& 10 & 8 & AdamW\\
    \midrule
    \multicolumn{7}{c}{Image Generation} \\
    \midrule
    Stable Diffusion v1.4 & 12 & 1e-4& 2e-4& 10000 iters& 1& AdamW \\
    Stable Diffusion v2.1 & 12 & 1e-4& 2e-4& 10000 iters& 1& AdamW \\
    Stable Diffusion v3 & 2 &1e-4 &5e-5 &10000 iters & 1& AdamW\\
    \midrule
    
    \end{tabular}
    \caption{Detailed hyper parameter settings of our experiments}
    \label{hyperparam}
\end{table*}

\subsection{Training hyper parameters}
Detailed training hyper parameters, including batch size, learning rate, threshold, optimizer, and epochs are summarized in Tab.~\ref{hyperparam}.

\end{document}